\documentclass[11pt]{article}
\usepackage{acl}
\usepackage{times}
\usepackage{latexsym}
\usepackage{placeins}

\usepackage[T1]{fontenc}
\usepackage[utf8]{inputenc}
\usepackage{tipa}          

\usepackage{microtype}
\usepackage{inconsolata}
\usepackage{graphicx}
\usepackage{amsmath}
\usepackage{tabularx}
\usepackage{float}
\usepackage{subcaption}
\usepackage{booktabs}
\usepackage{cuted}
\usepackage{capt-of}
\usepackage{enumitem}
\usepackage{xcolor}
\usepackage{comment}

\title{Adversarially Probing Cross-Family Sound Symbolism in 27 Languages}

\author{
  \textbf{Anika Sharma \quad Tianyi Niu \quad Emma Wrenn \quad Shashank Srivastava}  \\ [1ex]
  University of North Carolina at Chapel Hill \\ [1ex]
}

\begin{document}
\maketitle
\begin{abstract}
The phenomenon of sound symbolism, the non-arbitrary mapping between word sounds and meanings, has long been demonstrated through anecdotal experiments like Bouba–Kiki, but rarely tested at scale. We present the first computational cross-linguistic analysis of sound symbolism in the semantic domain of size. We compile a typologically broad dataset of 810 adjectives (27 languages $\times$ 30 items), each phonemically transcribed and validated with native-speaker audio. Using interpretable classifiers over bag-of-segment features, we find that phonological form predicts size semantics above chance even across unrelated languages, with both vowels and consonants contributing. To probe universality beyond genealogy, we train an adversarial scrubber that suppresses language identity while preserving size signal (also at family granularity). Language prediction averaged across languages and settings falls below chance while size prediction remains significantly above chance, indicating cross-family sound-symbolic bias. We release data, code, and diagnostic tools for future large-scale studies of iconicity.


\end{abstract}

\section{Introduction}
If you were watching a superhero movie called \textit{Lamonians vs. Grataks: The Phoneme Accords}, chances are you’d already be rooting for the Lamonians because they \textit{sound} like they would be nicer. In a 2009 \textit{Guardian} article, linguist David Crystal posed a thought experiment: when asked to judge two fictional alien races, most people instinctively sided with the Lamonians, drawn to the soft consonants (\textipa{/l/}, \textipa{/m/}, \textipa{/n/}) and long vowels and diphthongs that give the name its gentle, likable tone \cite{crystal2009guardian}. This phenomenon in which specific sounds systematically convey particular meanings is known as \textit{sound symbolism}, and it challenges the long-standing linguistic assumption that form and meaning are entirely arbitrary. But how can we be sure such intuitions reflect universal cognitive principles, rather than simply shared linguistic history? This paper introduces a method designed to answer this question.

Sound symbolism is most familiar in onomatopoeia, words like \textit{buzz} or \textit{crash} that imitate real-world sounds. It also manifests systematically across languages: in Yucatec Maya, vowel length signals event duration \citep{leguen_maya}; in Swedish, the prefix \textit{pj-} marks pejoration \citep{abelin1999soundsymbolism}; and in Japanese, consonants in food mimetics reflect perceived crispness \citep{raevskiy2023mimetics}. 

Despite evidence from individual languages, identifying cross-linguistic sound symbolism remains methodologically difficult because of two issues. First, identifying universal patterns is difficult because phonological similarities may reflect shared ancestry or areal contact rather than true sound symbolism. Second, phonological inventories differ: some languages lack certain sounds, which obscures potential effects. Yet, the question of cross-linguistic sound symbolism carries significance beyond theoretical linguistics. It may reveal cognitive universals in perception, improve cross-lingual transfer in low-resource NLP, and guide data-driven brand naming in commercial applications \citep{pathak2023luxury}.

In this work, we investigate whether sound symbolism for size holds across typologically diverse languages by testing if adjectives meaning “small” and “large” consistently share phonological features, regardless of language family. Our approach includes an adversarial setup designed to isolate potentially universal sound-symbolic patterns while controlling for language-specific influences. Our contributions are:
\begin{itemize}[leftmargin=*,itemsep=-2pt]
\item A cross-linguistic dataset of 800+ size adjectives (30 per language) from 27 languages from 13 language families, phonemically transcribed in IPA and validated through native speaker recordings to capture contrastive phonological distinctions.
\item An adversarial framework using gradient reversal \cite{ganin2015unsupervised} to suppress language-family signals while retaining semantic structure. The model maintains above-chance size classification (54.4\%) while reducing language identification to at chance (34.0\%) offering a new approach to disentangle universal patterns from genealogical relatedness.
\item Evidence that while vowels like /a/, /i/, and /o/ confirm traditional size-symbolism patterns, consonants—particularly voiced fricatives like \textipa{/\textrevglotstop/} and \textipa{/H/}—also contribute to size prediction across diverse language families, expanding beyond a purely vowel-centric account.
\end{itemize}

\section{Related Works}
\label{sec:related}

The classical view in linguistics considers the form--meaning link \emph{arbitrary} \citep{saussure1916}, with exceptions in \emph{iconicity} or \emph{sound symbolism}. Foundational experiments show that listeners map phonological form to perceived \emph{magnitude}; for example, \citet{sapir1929} found reliable size judgments for nonce forms differing only in vowel quality, and studies on maluma/takete (bouba/kiki) show cross-cultural mappings from sounds to rounded vs. angular shapes \citep{kohler1929,cwiek2022bouba}. Mechanistic accounts often rely on perceptual grounding, linking lower frequency (e.g., back/low vowels) to \emph{largeness} and higher frequency to \emph{smallness} \citep{ohala1994frequency}. Reviews can be found in \citet{dingemanse2016what} and \citet{sidhu2019names}. Recent surveys in language evolution and cognition note that iconic effects are typically \emph{modest} in size and \emph{domain specific} \citep{johansson2013deixis}.

\vspace{0.1cm}
\noindent \textbf{Magnitude symbolism}
Early studies on sound symbolism found that high front vowels (e.g., /i/) correlate with ``smallness'' whereas back/low vowels (e.g., /o, a/) correlate with ``largeness'' \citep{brown1955phonetic,jespersen1922symbolic,newman1933further,ultan1978size,berlin1994evidence}. The vowel pattern aligns with acoustic formants (lower F1/F2 for back/low vowels) and the frequency-code intuition \citep{ohala1994frequency}. However, consonantal contributions, including affricates, trills, and place/manner contrasts, also recur, e.g., consonants in North American diminutives \citep{nichols1971diminutive} and links between trilled /r/ and tactile ``roughness'' \citep{winter2022trilled}. Some English effects for size adjectives do not transfer to other parts of speech \citep{winter2021size}.

\vspace{0.1cm}
\noindent \textbf{Shape and other domains}
Beyond magnitude, iconic mappings span shape, texture, motion, affect, and social stance. Bouba--kiki (maluma--takete) patterns appear across languages and modalities \citep{cwiek2022bouba,sidhu2021maluma}, and recent work ties these effects more strongly to spectral properties than articulatory geometry \citep{passi2024bouba}. Broader surveys extend mappings to speed, brightness, and sentiment \citep{dingemanse2016what,sidhu2019names}.

\vspace{0.1cm}
\noindent \textbf{Cross-family regularities}
Extensive language and family-specific systems (e.g., Japanese mimetics \citep{hamano1986sound}, Dravidian and Austroasiatic expressives \citep{MBEmeneau,annamalai1968onomatopoeic,diffloth1972notes,diffloth1976expressives}, African ideophones \citep{childs1994african,childs2008research,samarin1965perspective}, and domain-specific English morphophonology \citep{bolinger1940word,rhodes1981athematic,calvillo2024sound}) demonstrate local structure. These systems also bring up two challenges for claims of universality: (1) \emph{genealogy/area} (areal diffusion and inheritance can induce clustering), and (2) \emph{lexical scope} (effects may depend on part of speech).

\vspace{0.1cm}
\noindent \textbf{Computational approaches \& genealogy control}
Computational studies quantifying form--meaning associations across families have typically found small but consistent biases \citep{blasi2016sound,gutierrez2016finding,pimentel2019meaning,pimentel2021finding,winter2022trilled,kilpatrick2023pokemon}. Some recent work shows that multimodal and language models can recover iconic patterns despite limited phonetic access \citep{alper2023kiki,loakman-etal-2024-ears}. 
Yet three methodological gaps remain. \emph{(1) Out-of-family validation:} many studies do not \emph{enforce} generalization to less-related languages. \emph{(2) Consonant coverage:} Consonants are often under-characterized relative to vowels. \emph{(3) Genealogical confounds:} most approaches rely on stratification or mixed-effects modeling \citep[e.g.,][]{blasi2016sound,pimentel2021finding}, but do not test whether size-predictive structure survives when language identity is suppressed. 
We use lexical/typological metrics (LDN/LDND) to build least-related partitions \citep{petroni2010ldn,bakker2009ldnd} and complement stratification with \emph{adversarial} representation learning to minimize genealogical effects~\citep{elazar2018adversarial,liu2020mitigating,chowdhury2021adversarial}.

\vspace{0.1cm}
\noindent \textbf{Scope and departure}
We specifically target these gaps with a controlled design, making effect sizes and cross-family signal directly comparable. We:

\vspace{0.1cm}
\begin{enumerate}[leftmargin=*,nosep]
    \item Assemble a curated, \emph{audio-validated}, multi-family dataset of size \emph{adjectives} to fix lexical scope while spanning typology;
    \item Evaluate \emph{out-of-family} generalization by training within similarity tertiles (via LDN/LDND) and testing on the held-out target language;
    \item Analyze \emph{both vowels and consonants}, identifying consonantal classes that contribute alongside classic vowel patterns;
    \item Introduce an \emph{adversarial probe} that suppresses language identity to clarify how much size signal remains when genealogy is minimized;
    \item Conduct targeted ablations removing different phonetic groups to assess their contributions.
\end{enumerate}

\section{Methodology}
\label{sec:approach}

In this section, we first describe our curated, audio-validated 27-language dataset with similarity tertiles, then describe our baseline classifiers, and then introduce an adversarial model that scrubs language identity while retaining size semantics. 

\subsection{Data Collection}

\textbf{Language Selection and Word list Compilation:} To investigate sound symbolism across typologically diverse languages, we compiled a dataset of 810 adjectives: 30 per language (15 denoting smallness, e.g., tiny, minuscule; 15 denoting largeness, e.g., huge, enormous)—spanning 27 languages from 13 language families. Languages were selected using the World Atlas of Language Structures (WALS) to maximize both genetic and areal diversity, while ensuring feasibility in terms of speaker access and resource availability.

Each adjective list was constructed by translating English seed words using DeepL \cite{DeepL2025}, Google Translate \cite{GoogleTranslate2025}, and bilingual dictionaries (\cite{abu1983hans},\cite{clines2018dictionary},\cite{kotey1998twi}). We manually filtered out borrowings and transliterations to prioritize native lexical items. Words were transcribed into the International Phonetic Alphabet (IPA) using phonemic transcription, which captures contrastive sound units while abstracting from fine-grained phonetic variation. Phonemic representations provide greater consistency across languages and reduce confounds from dialectal variation, particularly in languages such as Arabic and Spanish. We emphasize that our target is \emph{phonemic} structure rather than surface allophony.

\textbf{Phonemic Transcription Pipeline:} 
To ensure accuracy, we adopted a hybrid transcription pipeline. Where available, we used the XPF corpus --- a rule-based, linguistically grounded phoneme-mapping tool --- to generate initial IPA forms \citep{xpf2021}. We then collected audio recordings of the adjective lists from native speakers via language exchange platforms such as Tandem \cite{Tandem2025}. 

A trained phonetician reviewed the recordings using acoustic analysis tools and manually revised or reconstructed each transcription. In cases where no initial transcription was available, phonological forms were created from scratch based on the recordings. Domain experts, including language instructors and researchers, also reviewed or annotated the data for several languages to ensure quality and accuracy. In Appendix \ref{app:phon_cov} we report the phoneme coverage  for each language. 

\noindent \textbf{Genealogical Distance Measures:} To control for shared linguistic ancestry, we used a precomputed dataset of normalized Levenshtein distances between languages, compiled by \citet{nielsen2017} following the method introduced by \citet{bakker2009}. These distances were computed from phonologically transcribed 40-word Swadesh lists, standardized core vocabulary chosen for cross-linguistic comparability. The values reflect average phonological edit distance between word pairs across languages and were used to group languages into similarity bins.

\begin{figure*}[h]
    \centering
    \includegraphics[width=1.0 \linewidth]{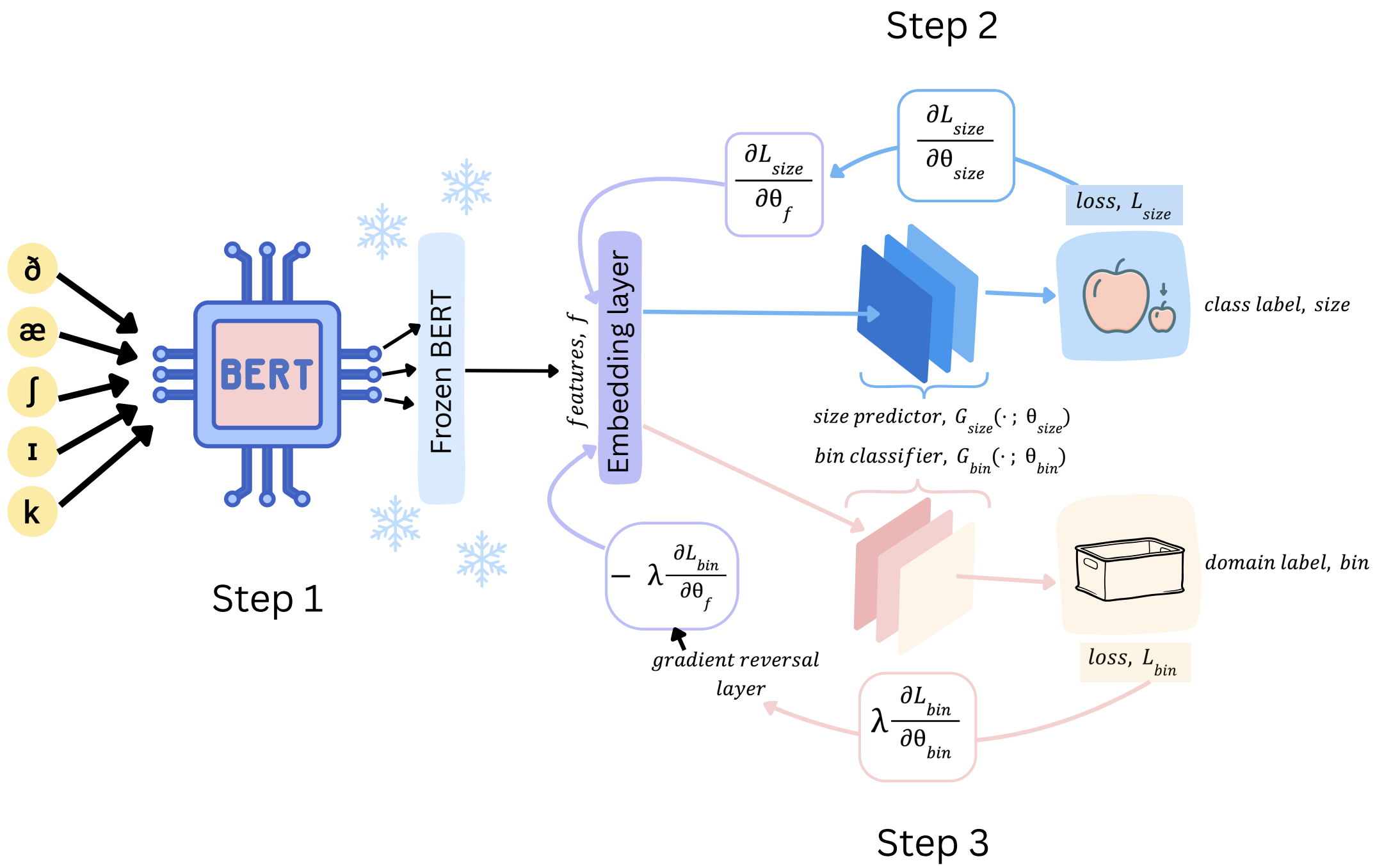}
    \caption{Overview of our adversarial training pipeline. Stage 1: pretraining on IPA-transcribed words from WikiPron and freezing BERT as a fixed feature extractor. Stage 2 (Forward): the features feed both the Size Classifier ($L_\text{size}$) and the Language-bin Adversary ($L_\text{bin}$), with the GRL (gradient reversal layer) acting as an identity function. Stage 3 (Backward): implement the adversarial training. The GRL reverses the domain gradient, causing the feature representations to update against the adversary (maximizing $L_\text{bin}$) while still supporting the primary goal of size classification (minimizing $L_\text{size}$).} 
    \label{fig:adversarial-setup}
\end{figure*}

\subsection{Baseline Classifiers}
First, we trained logistic regression and decision tree classifiers to predict size. Each word was represented as a bag-of-phoneme vector based on IPA symbol counts. Because standard tokenization mishandles diacritics and other IPA symbols, we used the \textit{ipatok} \cite{pavelsof2025pavelsof} library for phoneme segmentation. We excluded suprasegmentals (tone, stress, intonation) due to their language-specific encoding and the lack of expert phoneticians native to each language.

We used a leave-one-language-out design. For each of the 27 languages, we held it out as the test set and trained on the remaining 26 languages. For each target language $\ell$, we rank the remaining languages by \emph{increasing} similarity (LDN) to $\ell$ and partition into tertiles: \emph{Most Similar}, \emph{Somewhat Similar}, and \emph{Least Similar}. For each tertile (bin) $t$, we train exclusively on languages in $t$ and evaluate on $\ell$, averaging results over all $\ell$ to obtain bin-wise means and 95\% CIs. Decision trees used maximum depth 6 and minimum samples per split of 2–10 to prevent overfitting. Accuracy was the primary evaluation metric. 

To analyze feature importance, we extracted logistic regression coefficients to identify phonemes most strongly associated with smallness or largeness. For decision trees, we identified the phoneme at the root split (highest Gini impurity reduction) for each target language. We then compared these across languages to see which phonemes were the primary discriminators.

\subsection{Adversarial Scrubber}

We combine phonetic pretraining with adversarial learning to suppress language-family signals while preserving size-symbolic information. Given the modest scale of our curated dataset (810 adjectives across 27 languages), we pretrain a compact BERT encoder on 1.6 million IPA-transcribed words from WikiPron \cite{lee-etal-2020-massively}, an open-source tool for retrieving IPA transcriptions from Wiktionary. Using masked language modeling, this pretraining allows the encoder to build general phonotactic and articulatory knowledge across typologically diverse languages, providing a stronger basis for identifying symbolic patterns in our downstream task. The encoder consists of 2 transformer layers with 128 hidden units and a vocabulary of 115 IPA characters. After 2 epochs, the encoder is frozen and used as a fixed feature extractor.

We then learn a 64-dimensional projection and two task heads: size (binary) and similarity bin label (multi-class). We implement the adversary with a gradient-reversal layer in the style of domain-adversarial training \citep{ganin2015unsupervised}. We fine-tune a linear embedding layer, a size classifier, and a bin adversary on our curated dataset, where each word is annotated with both a binary size label and a similarity bin label. The model architecture includes the following components:

\begin{itemize}[leftmargin=*,noitemsep]
    \item \textbf{Encoder (E):} A single linear layer (128→64) with 
    dropout that projects frozen BERT outputs into a shared 
    representation space. Receives adversarial gradients from the bin 
    classifier.
    \item \textbf{Sound Symbolism Classifier (C):} A two-layer feed-forward network with ReLU and dropout, then a softmax ofver two classes (small, large).
    \item \textbf{Bin Adversary (A):} A two-layer feed-forward network with ReLU and dropout, followed by softmax over 3 similarity bins. Connected to the encoder via a gradient reversal layer that multiplies gradients by $-\lambda$ during backpropagation.
\end{itemize}

To encourage the encoder to discard similarity bin information (a proxy for genealogical relatedness) while preserving size-predictive structure, we optimize a minimax objective:

\vspace{-0.3cm}
\begin{align}
\min_{E, C} \ \max_{A} \quad
& \mathcal{L}_{\text{size}}(C(E(x)),\ y_{\text{size}}) \nonumber \\
& -\ \lambda \cdot \mathcal{L}_{\text{bin}}(A(E(x)),\ y_{\text{bin}})
\end{align}

Here, $\mathcal{L}_{\text{size}}$ and $\mathcal{L}_{\text{bin}}$ are cross-entropy losses for size and bin classification, respectively. We use a leave-one-language-out design: for each of the 27 languages, we hold it out as the test set and train on the remaining 26 languages. Each training language is assigned to one of 
three similarity bins (most/somewhat/least similar) relative to the held-out target language, like what was done for the baseline classifiers. $\lambda$ is ramped from 0 to 1 over 20 epochs following the schedule in \citet{ganin2015unsupervised}:

\begin{equation}
\lambda(p) = \lambda_{\text{max}} \cdot \left(\frac{2}{1 + \exp(-10p)} - 1\right),
\end{equation}
where $p = \text{epoch}/\text{total epochs}$ and $\lambda_{\text{max}} = 1.0$.

We implement this using a gradient reversal layer (GRL) 
\citep{ganin2015unsupervised}. During the forward pass, both the size classifier and bin classifier receive the same encoder output. During the backward pass, the GRL reverses gradients from the bin classifier (multiplying by $-\lambda$) before they reach the encoder. This means: 
(1) the size classifier updates the encoder to improve size prediction, while (2) the bin classifier updates the encoder to \emph{worsen} bin prediction. The result is an encoder that learns representations informative for size but uninformative for genealogical similarity bins.

We apply dropout ($p=0.1$) before classification. Chance levels are 50\% for size and 33.3\% for bin classification ($K=3$ bins). Our evaluation focuses on two key metrics: (1) test size accuracy on the held-out target 
language (should remain above chance with adversarial training), and (2) train bin accuracy across the 26 training languages (at chance indicates successful suppression). 

To verify that the model is able to learn both bin and size information, we train a baseline with $\lambda=0$ (no adversarial pressure) for comparison.

\section{Results}
\label{sec:results}

\begin{figure*}[h]
    \centering
    \includegraphics[width=1.0 \linewidth]{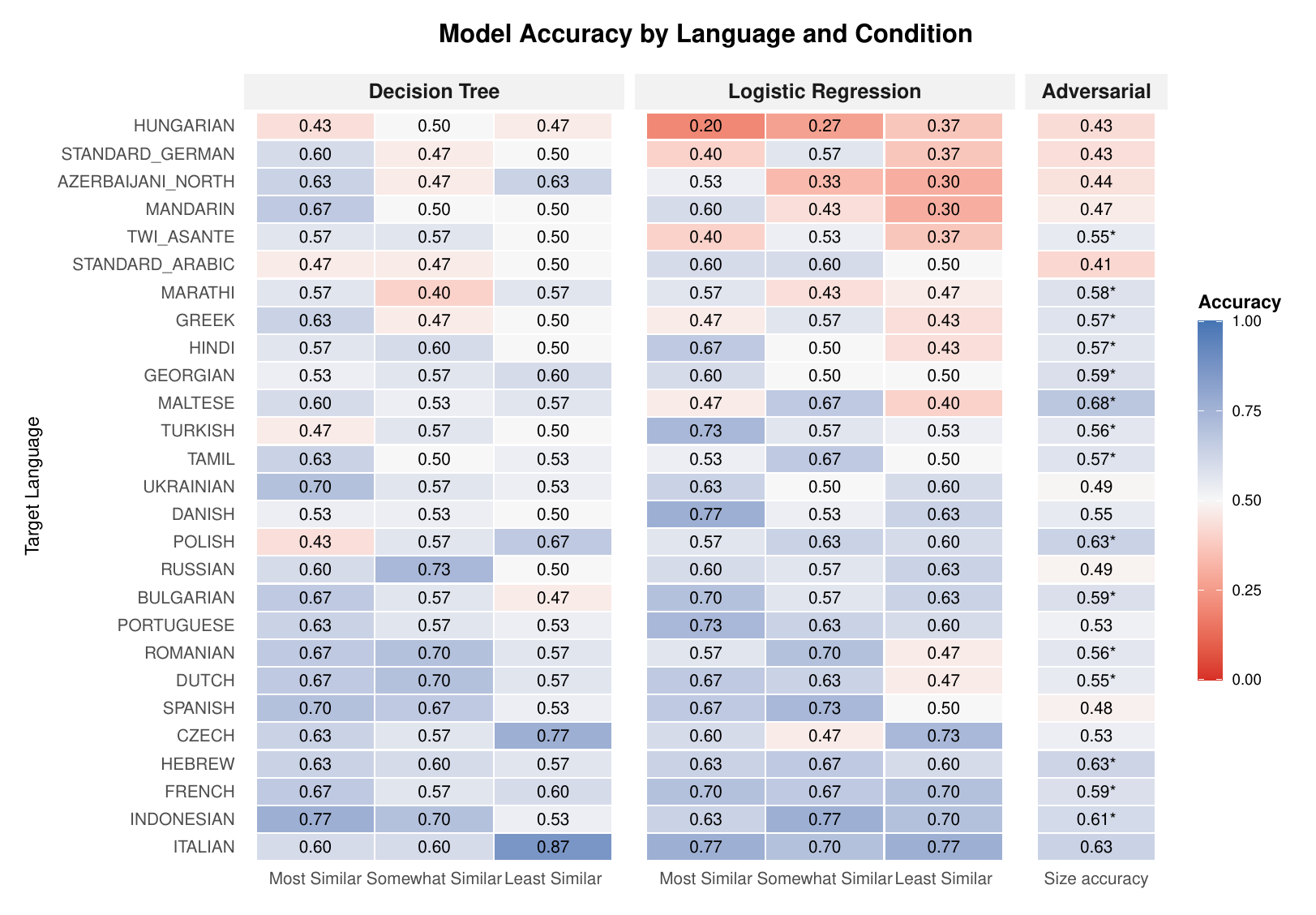}
    \caption{Baseline classifiers perform above chance for most languages, but accuracy becomes unstable with increasing linguistic dissimilarity—several languages drop to or below chance in the Least Similar bin. The adversarial model suppresses genealogical cues while maintaining above-chance size prediction in 15 of 27 languages (asterisks). While baselines achieve higher average accuracy, the adversarial setup tells us which languages show size–sound patterns that aren't just artifacts of language family.} 
    \label{fig:heatmap-languages}
\end{figure*}

\begin{table}[h]
\centering
\small
\renewcommand{\arraystretch}{1.1}
\setlength{\tabcolsep}{4pt}
\begin{tabularx}{\columnwidth}{lXXX}
\toprule
\textbf{Model} & \textbf{Most Similar} & \textbf{Somewhat Similar} & \textbf{Least Similar} \\
\midrule
Logistic Reg. & 59.2\textsuperscript{***} & 57.0\textsuperscript{**} & 52.2 \\
Decision Tree & 60.2\textsuperscript{***} & 56.4\textsuperscript{***} & 55.8\textsuperscript{**} \\
\bottomrule
\end{tabularx}
\caption{Mean accuracy (\%) by model and similarity bin. 
Asterisks (***) indicate bins where classifiers performed significantly above chance ($p < 0.001$). \textcolor{blue}. Results are averaged over 27 languages for each bin. Chance accuracy is 50.0\%.}
\label{tab:baselineAccuracies}
\end{table}

\subsection{Baseline classification accuracy}

\textbf{Although model performance decreases with linguistic dissimilarity, classifiers maintain above-chance accuracy on average}. Decision Trees performed significantly above chance across all three similarity bins (p < 0.01), with large to medium effect sizes (Cohen's $d$ = 1.23, 0.78, 0.65 for the \textit{Most similar}, \textit{Somewhat similar}, and \textit{Least similar} bins, respectively). Logistic Regression classifiers also performed significantly above chance for the \textit{Most similar} and \textit{Somewhat similar} bins (Cohen's $d$ = 0.73 and 0.60; $p < 0.01$ for both), but not for the least similar bin, which approached chance-level accuracy. Accuracy generally decreased going from the \textit{Most Similar} bin to the \textit{Least Similar} bin (Figure~\ref{fig:heatmap-languages}) for most languages. \\

\vspace{0.2cm}

\begin{figure}[!t]
  \centering
  \includegraphics[width=\columnwidth]{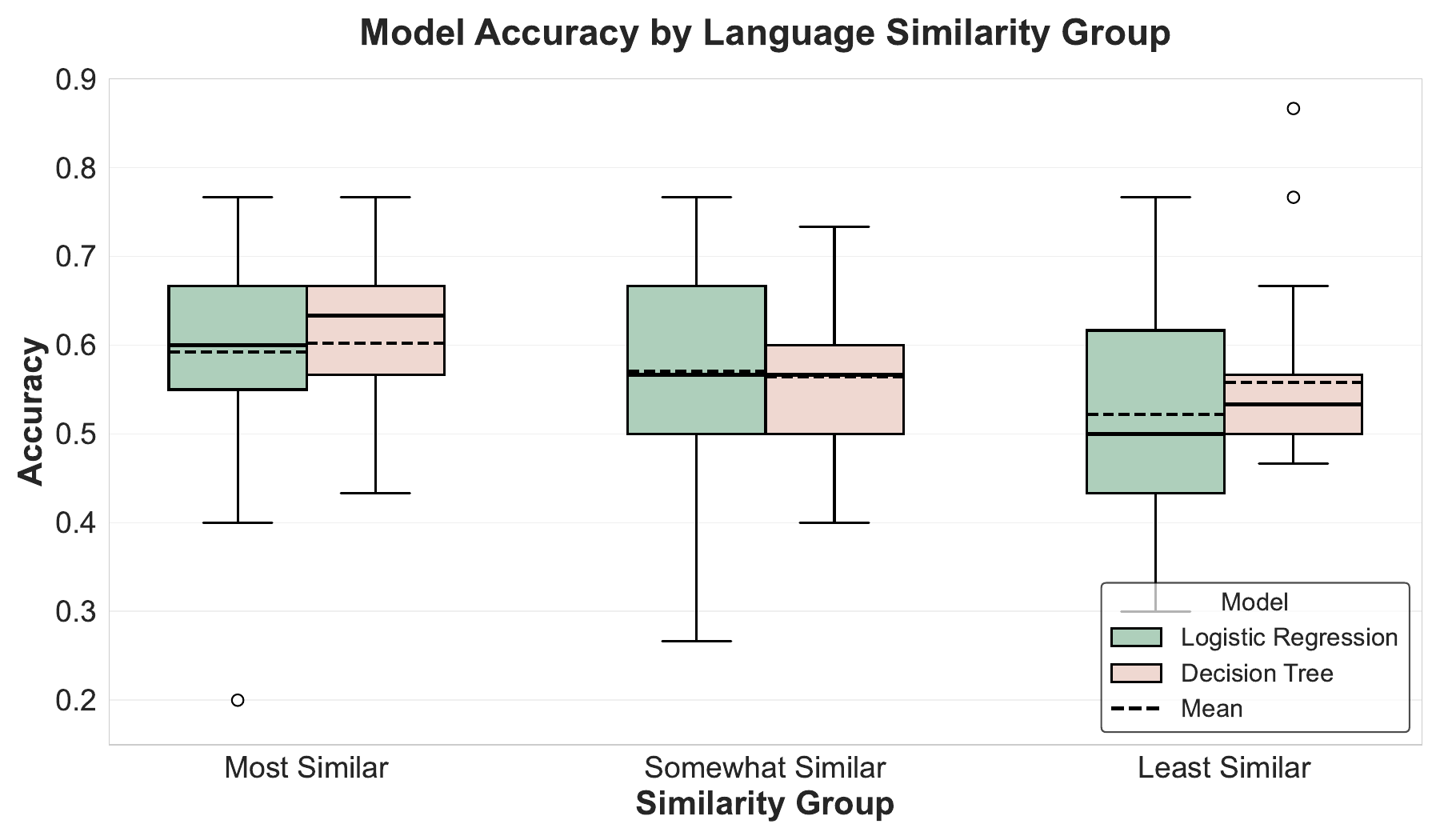}
  \caption{Both baseline classifiers had above chance accuracies for size classification across all bins (Most similar, Somewhat Similar and Least similar). However, accuracy did descrease with bin dissimilarity.}
  \label{fig:classifierAcc}
\end{figure}

\textbf{Differences in accuracy across similarity bins were not statistically significant for both baseline classifiers:}
 Figure~\ref{fig:classifierAcc} shows the distribution of model performance across similarity bins. Detailed language accuracies are in ~\ref{fig:heatmap-languages}. In general, we note that accuracy declines with increasing dissimilarity for both Decision Trees and Logistic Regression. However, differences in accuracy across similarity bins were not found statistically significant. A one-way ANOVA showed that differences in model performance across bins were not statistically significant ($p = 0.08$ for Logistic Regression; $p = 0.19$ for Decision Trees). We note that this does not mean that there is no trend, but that we do not conclusively find it with our sample sizes.

\subsection{Phonological Features \& Size Symbolism}

\textbf{For Decision Trees, vowels \textipa{/a/} and \textipa{/i/} are the most common root splits:} We looked at the root split of each Decision Tree—the feature that best separates big from small words at the top of the tree. Across 81 language–bin combinations, vowels dominated: \textipa{/a/} appeared in 22 cases (27\%) and \textipa{/i/} in 19 (24\%). Consonants were less common, with fricatives (\textipa{/h/}: 5, \textipa{/v/}: 4, \textipa{/H/}: 3, \textipa{/s/}: 3), nasals (\textipa{/m/}: 4), and stops (\textipa{/k/}: 3) appearing in a minority of models. The dominance of \textipa{/a/} and \textipa{/i/} (low and high vowels respectively) aligns with cross-linguistic evidence that vowel height is a key correlate of size symbolism. This pattern held across bins: \textipa{/i/} and \textipa{/a/} accounted for 63\% of root splits in the Most Similar bin and 55\% in the Least Similar bin, though more varied features like \textipa{/Q/} emerged in the latter. The Somewhat Similar bin showed the most variation, with consonants like \textipa{/g/} and \textipa{/h/} as root splits in several models.

Models with \textipa{/a/} as the root split performed above chance in 16/22 cases, with mean accuracy significantly exceeding chance (one-sample $t$-test; $M = 0.57$, $t(21) = 3.38$, $p < 0.01$). Similarly, models with \textipa{/i/} performed above chance in 15/19 cases ($M = 0.58$, $t(18) = 4.52$, $p < 0.001$).  These patterns occurred across diverse language families: \textipa{/a/} was the top split for Turkish (Turkic), Mandarin (Sino-Tibetan), Hebrew (Semitic), Italian (Romance), and Polish (Slavic), while \textipa{/i/} appeared for Marathi (Indo-Aryan), Spanish (Romance), Greek (Hellenic), and Hungarian (Uralic).

\textbf{For Logistic Regression, \textipa{/o/} is the strongest predictor of largeness and \textipa{/i/} of smallness.} For Logistic Regression, we first filtered for phonemes with consistent coefficient direction across models (>60\% same sign), then computed mean coefficients for those phonemes. This avoids averaging across phonemes that flip directions between languages. Phonemes with the largest mean coefficients were language-specific variants that appeared in only a subset of models. For largeness, \textipa{/H/} had a mean coefficient of +1.32 across the 37 models where it appeared, and \textipa{/\~{e}/} had +1.12 across 26 models. For smallness, \textipa{/\~{a}/} had a mean coefficient of -0.86 across 26 models and \textipa{/n\super{j}/} had a mean coefficient of -0.85 across 35 models. These phonemes exist only in certain languages, so their high coefficients likely reflect phonotactic patterns within language families.

Among vowels present in all 81 models, \textipa{/o/} showed the strongest association with largeness (100\% positive, mean = +0.67), while \textipa{/i/} showed the strongest association with smallness (99\% negative, mean = -0.44). The vowel \textipa{/a/} (the most frequent root split for Decision Trees) had a smaller effect/mean coefficient (+0.26) but still pointed consistently toward largeness.

Both baseline classifiers agreed on \textipa{/i/} as a key feature for smallness. For largeness, Decision Trees favored \textipa{/a/} as a root split while Logistic Regression assigned stronger coefficients to \textipa{/o/}. However, both \textipa{/o/} and \textipa{/a/} are back/low vowels which is consistent with classic findings linking vowel height to size perception.

\subsection{Adversarial Training Results}

\begin{figure}[t]
    \centering
    \includegraphics[width=1.0\linewidth]{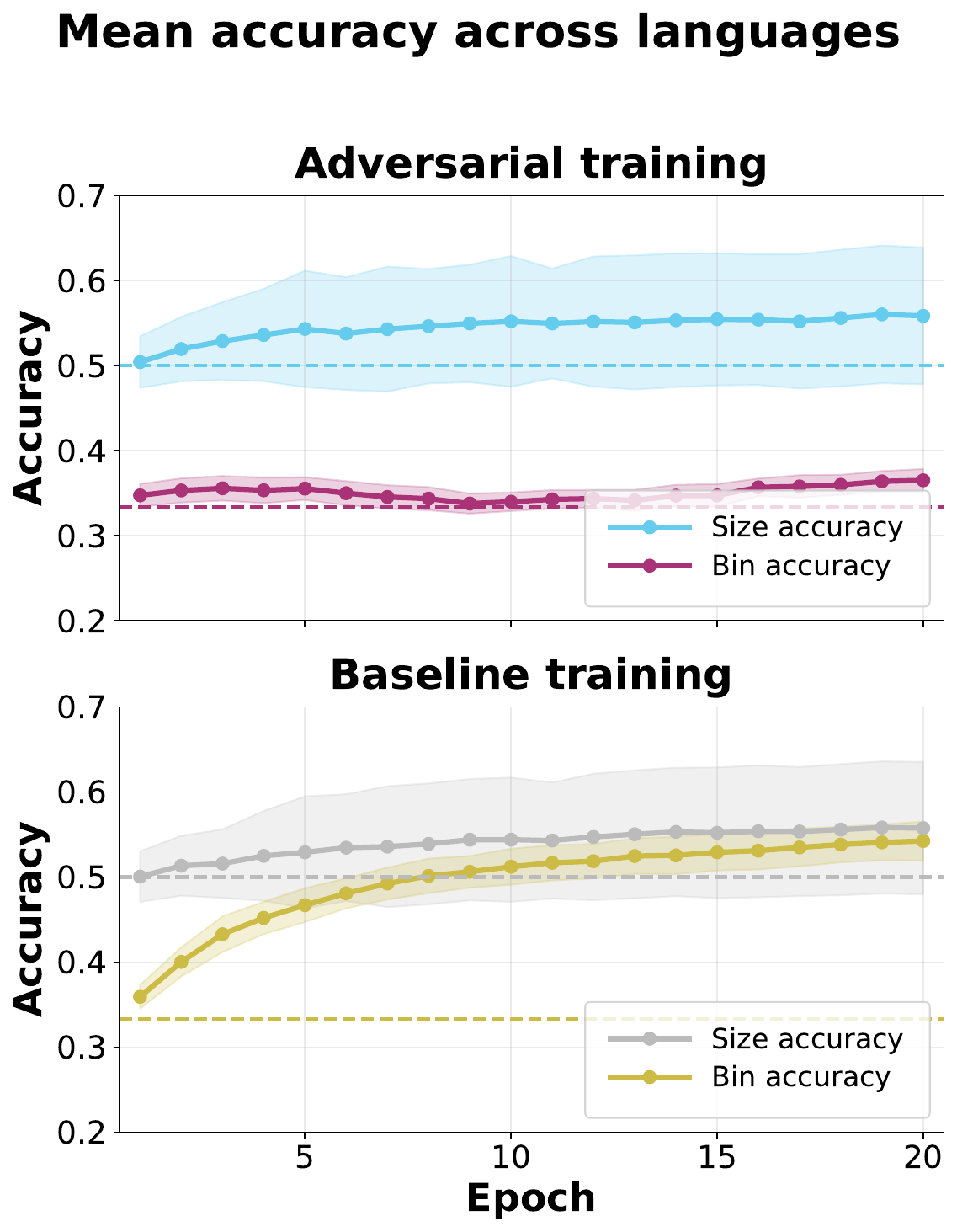}
    \caption{Average test size accuracy and bin accuracy across 27 languages over 20 epochs. Without adversarial pressure, bin accuracy steadily increases, indicating that the model is using language-specific cues. Under adversarial training, bin accuracy is reduced to chance level, while size accuracy remains consistently above chance and is similar to the baseline model. This suggests that the adversarial objective limits the use of language cues without substantially affecting performance on the size prediction task.}
    \label{fig:adv_training_2}
\end{figure}

 We trained two model variants for each of the 27 target languages: (1) adversarial ($\lambda$ ramped from 0 to 1), and (2) baseline ($\lambda=0$, no adversarial pressure). For each variant, we conducted 5 runs. 

\textbf{Epoch selection.} For the adversarial condition, we selected the epoch (greater than or equal to 8, when $\lambda \approx 1$) that maximized the difference between train size accuracy and train bin accuracy, with train bin accuracy at or above chance (33.3\%; if no such epoch existed, we removed this constraint). This identifies the point where bin suppression is strongest while size information is preserved. For the baseline ($\lambda=0$), we selected the epoch with highest train bin accuracy. For the baseline ($\lambda=0$), we selected the epoch (greater than 10) with highest train bin accuracy.

\textbf{Adversarial training successfully suppressed language identification while maintaining size classification accuracy.} The adversarial model achieved test size accuracy of 54.4\% and train bin accuracy of 34.0\% (at chance), compared to the baseline's 55.4\% and 54.4\%. A paired $t$-test across languages found no significant difference in size accuracy between conditions, $t(26) = 1.76$, $p = .090$, confirmed that suppressing language identity did not significantly affect size prediction.

For each language, we tested significance across 5 runs using one-sample $t$-tests. Size accuracy was tested against chance (50\%) using a one-tailed test ($p < 0.05$). Bin accuracy was tested against chance (33.3\%) using a two-tailed test, where successful suppression corresponds to non-significance. We report bin suppression results at two thresholds ($p \geq 0.01$ and $p \geq 0.05$). Two languages (Mandarin, Ukrainian) showed no variation across runs, preventing statistical testing.

Fifteen of the remaining 25 languages (60\%) showed above-chance size accuracy ($p < 0.05$) while bin accuracy remained at chance ($p \geq 0.01$): Bulgarian, Dutch, French, Georgian, Greek, Hebrew, Hindi, Indonesian, Maltese, Marathi, Polish, Romanian, Tamil, Turkish, and Twi Asante. Under a stricter bin threshold ($p \geq 0.05$), 6 languages met both criteria: Dutch, Greek, Hindi, Indonesian, Marathi, and Twi Asante.

\begin{table*}[t]
\centering
\small
\renewcommand{\arraystretch}{1.2}
\setlength{\tabcolsep}{5pt}
\begin{tabularx}{\textwidth}{lXccc}
\toprule
\textbf{ID} & \textbf{Ablation Condition} & \textbf{LR Mean} & \textbf{DT Mean} & \textbf{LR / DT by Bin} \\
\midrule
0 & Baseline (all phonemes) & 56.1 & 57.5 & 59.2 / 57.0 / 52.2 \quad  | \quad  60.2 / 56.4 / 55.8 \\
1 & Scrambled Labels & 48.8 & 50.4 & 48.6 / 47.9 / 50.0 \quad  | \quad  50.8 / 50.0 / 50.4 \\
2 & No Vowels & 56.1 & 54.2 & 58.6 / 52.8 / 51.1 \quad  | \quad  55.8 / 54.6 / 52.2 \\
3 & High-Frequency Phonemes (see Appendix~\ref{app:phonemeLists}) & 57.9 & 62.1 & 60.4 / 54.7 / 57.9 \quad  | \quad  67.3 / 60.6 / 62.1 \\
4 & Only Plosives (see Appendix~\ref{app:phonemeLists}) & 54.5 & 56.4 & 56.7 / 53.5 / 53.2 \quad  | \quad  58.6 / 55.9 / 54.6 \\
5 & Only Nasals (see Appendix~\ref{app:phonemeLists}) & 51.2 & 54.2 & 52.0 / 51.0 / 51.8 \quad | \quad 53.0 / 51.6 / 53.0 \\
\bottomrule
\end{tabularx}
\caption{Mean and bin-wise accuracy (\%) for logistic regression (LR) and decision tree (DT) classifiers across ablation conditions. Bin results are shown in the format: Most / Somewhat / Least Similar. Full phoneme sets are listed in \hyperref[app:phonemeLists]{Appendix~\ref*{app:phonemeLists}}.}
\label{tab:ablationResults}
\end{table*}
\subsection{Ablation Experiments}
\label{sec:ablations}
Table~\ref{tab:ablationResults} summarizes different ablation analyses for the baseline classifiers. 
\begin{enumerate}[leftmargin=*,itemsep=-2pt]
    \item \textbf{Scrambled Size Labels:} For each language, we randomly shuffled test set size labels while keeping class balance. This checked if models truly captured phoneme–meaning associations instead of memorizing artifacts, expecting chance performance if semantics were random. Logistic Regression accuracies did not exceed chance in any bin.  Average accuracies were 48.6\% for the 'most similar' group (p = 0.80), 47.9\% for 'somewhat similar' (p = 0.90), and 50.0\% for 'least similar' (p = 0.50). Similarly, for decision trees, all mean accuracies were statistically non-significant. The 'most similar' bin achieved 50.9\% accuracy (p = 0.267), the 'somewhat similar' bin was at 50.0\% (p = 0.500), and the 'least similar' bin was at 50.4\% (p = 0.416).
    \item \textbf{No vowels:} Given the importance of vowels for size classification in both the Logistic Regression and Decision Tree baselines, the goal of this ablation was to examine how much of the size-sound symbolism signal remains once all vowels are removed from the input. Logistic regression stayed above chance when trained on the most similar languages (58.6\%), but accuracy dropped for the somewhat similar and least similar bins (52.8\% and 51.1\%). Decision trees showed the same pattern (55.8\%, 54.6\%, 52.2\%). These results suggest that consonants do contribute to size distinctions, but their effect weakens as languages become less related.
    \item \textbf{High Frequency Phones only:}  This ablation tested whether performance would change when models are restricted to only the most common phonemes cross-linguistically. The list of 23 segments was provided by our collaborating phonetician and includes phones that are widely attested across the world’s languages. Both classifiers performed nearly as well as the full model. This suggests that much of the sound-symbolic signal can be recovered using only the most typologically frequent phonemes, and that rarer or language-specific phonemes may not be necessary to detect size-related patterns.
    \item \textbf{Stops vs. Nasals (consonants):} We compared stops and nasals, chosen for cross-linguistic frequency and articulatory contrast (stops involve closure and bursts: nasals use continuous nasal airflow). Stops outperformed nasals in both models (54.5\% vs. 51.2\% for logistic regression; 56.4\% vs. 54.2\% for decision trees).
    \end{enumerate}
\section{Discussion}
\noindent \textbf{Interpreting Phonetic Trends}
Previous studies link back vowels to largeness due to larger sub-oral cavities \citet{shinohara2010cross}, while front vowels like \textipa{/i/} are tied to smallness via higher formant frequencies and smaller oral apertures (Sapir, 1929). Our findings from the baseline classifier feature importance affirm this pattern, as back/central vowels consistently predict largeness, and front vowels signify smallness. The importance of these vowels in sound symbolism is further substantiated by our "no vowels" ablation study, which suggests that vowel-based symbolism is more perceptually grounded than consonant-based cues. Consonants alone predicted size only for the most similar bin and dropped toward chance for the somewhat and least similar bins, consistent with vowels driving the cross-family signal.

Similarly, voiced consonants have been shown to evoke larger images than voiceless ones \cite{newman1933further,shinohara2010cross}, as our results confirm. The frequency code hypothesis \cite{ohala2006frequency} suggests low-frequency acoustic energy signals largeness due to resonance frequency's inverse correlation with cavity size. This may explain why voiced fricatives like \textipa{/ \textrevglotstop /} and \textipa{/H/} (characterized by substantial low-frequency energy and maximal vocal tract expansion) were highly predictive features for certain languages.

\vspace{0.1cm}
\noindent \textbf{Interpreting the Adversarial Results}

Both baseline and adversarial classifiers performed above chance on size prediction. The languages meeting both criteria --- above-chance size accuracy and at-chance bin accuracy --- belonged to a diverse range of language families and covered 7 of the 13 families in our dataset. However, generalizing to universal sound symbolism requires caution, as even genealogically distant languages may share deeper phylogenetic connections. With the adversary active, language identification dropped to chance while size classification remained stable. A paired t-test confirmed no significant difference between adversarial and baseline size accuracy. This suggests that size-related phonetic cues persist even when language identity is suppressed.

The small performance gap between models reveals both the strength and limits of the observed sound symbolism. The signal persists under language suppression but may be relatively shallow. These results suggest that certain phonetic trends (vowel backness and voicing) may carry size associations across genealogically diverse languages, though not necessarily constituting a universal mapping. Confirming this would require testing on larger, more representative data with finer-grained language distinctions than Levenshtein binning. 

\section{Conclusion}
\label{sec:conclusion}

Our work here recasts size symbolism as a measurement problem: what signal remains after rigorous controls? Using interpretable models and an adversarial probe, we see a consistent cross-family signal in which consonants as well as vowels contribute. Whilst our findings were consistent with vowel dominant sound symbolism theories, other classes like voiced fricatives also proved highly predictive, suggesting acoustic properties like frequency may better explain cross-linguistic patterns than vowel-centric articulatory theories. Future work can expand typological coverage and lexical categories, model suprasegmentals/diacritics more fully, incorporate acoustic/perceptual evidence, and move from binary labels to continuous size scales. Replicating our approach of reporting least-similar bin performance and adversarial probing can help better accumulate evidence across studies. In this sense, our contribution here is a measurement design that can extend to other domains of iconicity.

\section*{Limitations}
A key limitation of this study is the modest size of our dataset, which reflects the practical difficulty of conducting cross-linguistic research at scale. Recruiting native speakers for 27 languages, particularly those with limited academic or online presence, posed a significant logistical challenge. Ideally, all transcriptions would be vetted by native speakers with formal linguistic training, but such individuals are extremely rare for many of the languages in our sample. While we aimed for genealogical diversity using WALS and intentionally included languages from 13 families, the dataset remains slightly skewed toward Indo-European languages due to the relative accessibility of speakers. Additionally, we relied on phonemic rather than phonetic transcriptions; although phonemic representations offer greater cross-speaker consistency, they may miss subtle sound-symbolic cues present in actual pronunciation. 

Methodologically, our adversarial probe is specific to a model class and can under-or over-scrub structure correlated with language identity. In our analysis, suprasegmentals are only partially incorporated; some languages (e.g., tone systems) might merit full prosodic modeling. Finally, larger samples covering a broader range of languages, as well as perceptual experiments may be needed to fully judge sound universals.

\section*{Acknowledgments}
We thank Christopher Lucas (SOAS, University of London), Michael Flier (Harvard University),  Katya Pertsova (UNC Chapel Hill), Julie De Lataulade de Laas (UNC Chapel Hill), Paul Roberge (UNC Chapel Hill) and Nina Topintzi (Aristotle University of Thessaloniki) for their expertise in providing feedback on our word lists and IPA transcriptions for Maltese, Ukrainian, Russian, French, German, and Greek.

\bibliography{custom}

\appendix
\clearpage
\onecolumn

\section{Detailed baseline training plots by language (adversary off)}
\label{app:baseline_lang}
\begin{figure}[!htbp]
  \centering
  \includegraphics[width=\linewidth,height=0.85\textheight,keepaspectratio]{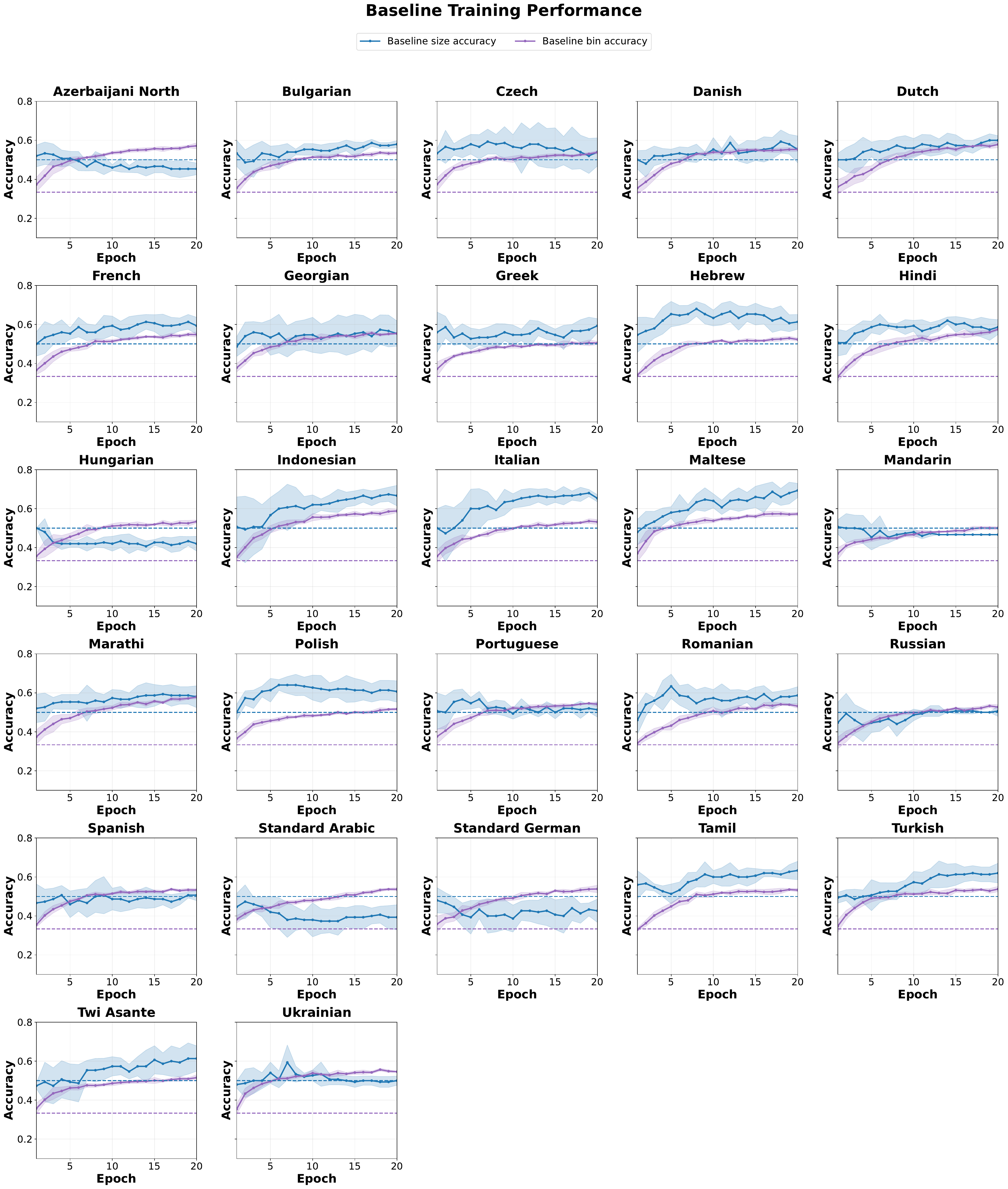}
  \caption{Baseline training dynamics ($\lambda=0$) for each of the 27 languages. Without adversarial pressure, bin accuracy rises above chance, indicating the model learns language-specific cues. Shaded regions indicate $\pm1$ SD across 5 runs.}
  \label{fig:baseline}
\end{figure}

\clearpage
\section{Detailed adversarial training plots by language}
\label{app:adv_lang}
\begin{figure}[!htbp]
  \centering
  \includegraphics[width=\linewidth,height=0.85\textheight,keepaspectratio]{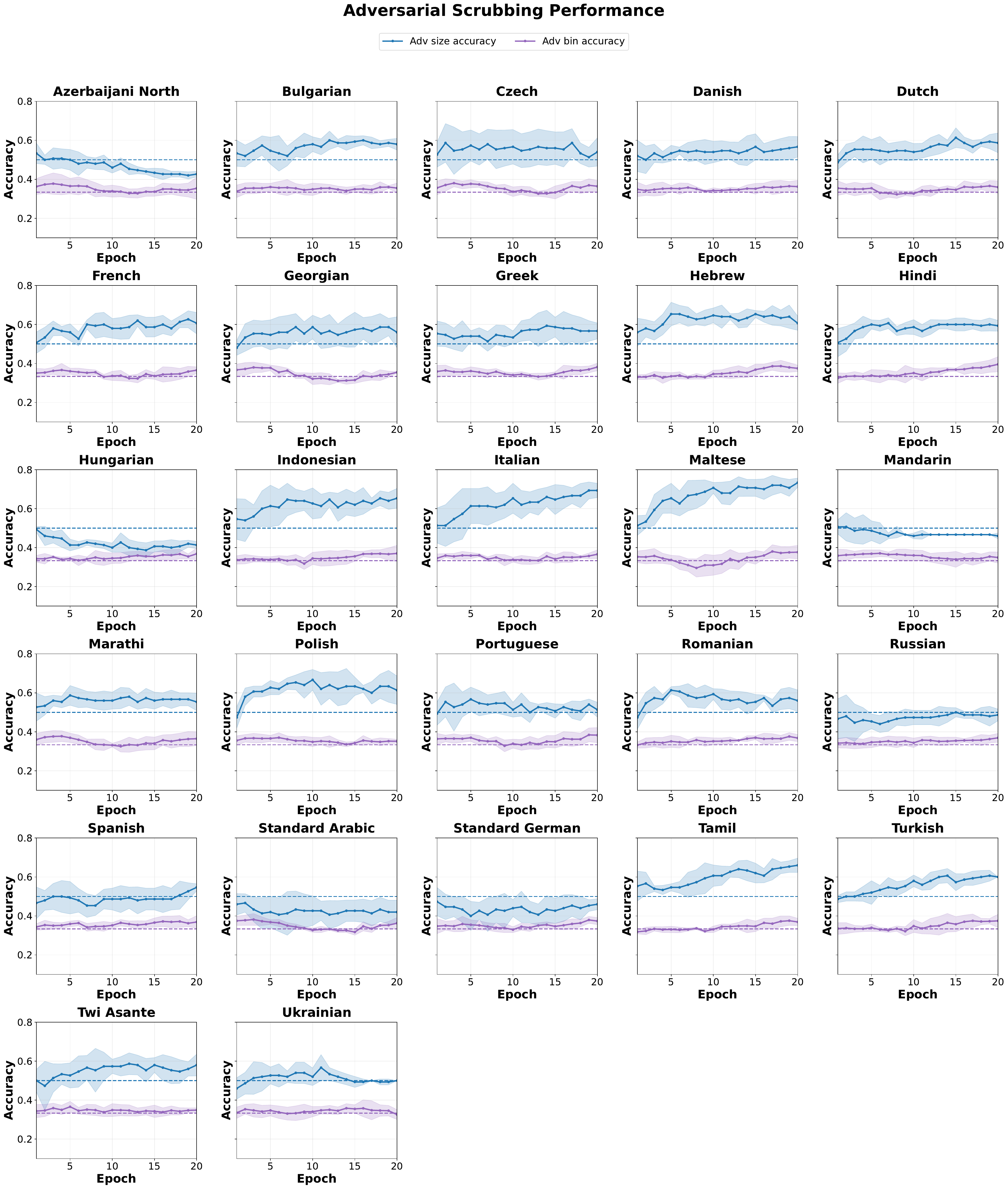}
  \caption{Adversarial training dynamics for each of the 27 languages. Shaded regions indicate $\pm1$ SD across 5 runs.}
  \label{fig:adv}
\end{figure}

\clearpage

\section{Ablation Phone Sets}
\label{app:phonemeLists}

\begin{figure}[htbp]
    \centering
    \includegraphics[width=\textwidth]{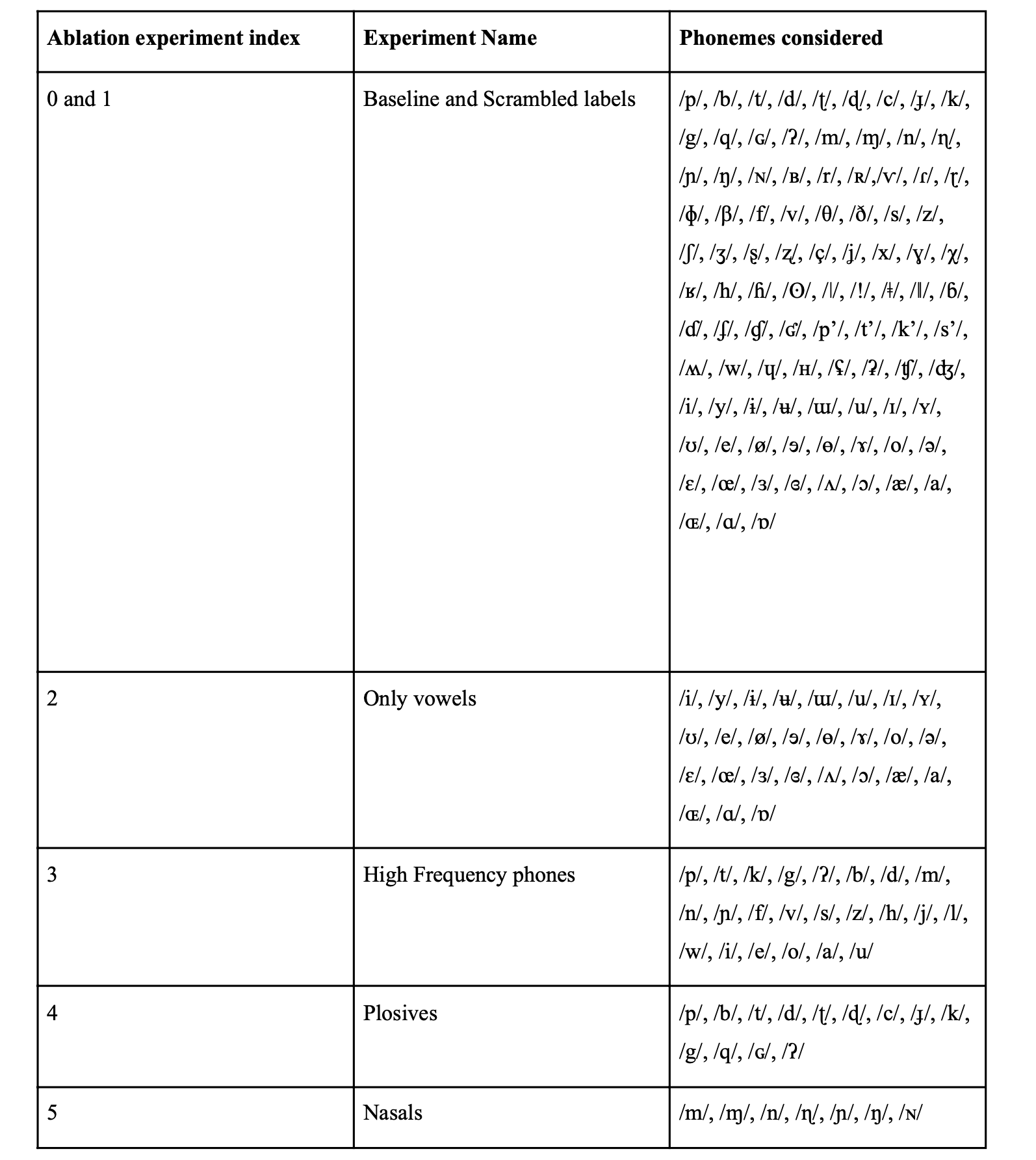}
    \label{fig:appendix_chart}
\end{figure}

\clearpage
\section{Token Coverage by Language}
\label{app:phon_cov}
\noindent Token coverage statistics for each language, showing the proportion of phonemes in our dataset covered by the BERT vocabulary trained on WikiPron.\par\medskip

\centering
\includegraphics[width=\textwidth, height=0.85\textheight, keepaspectratio]{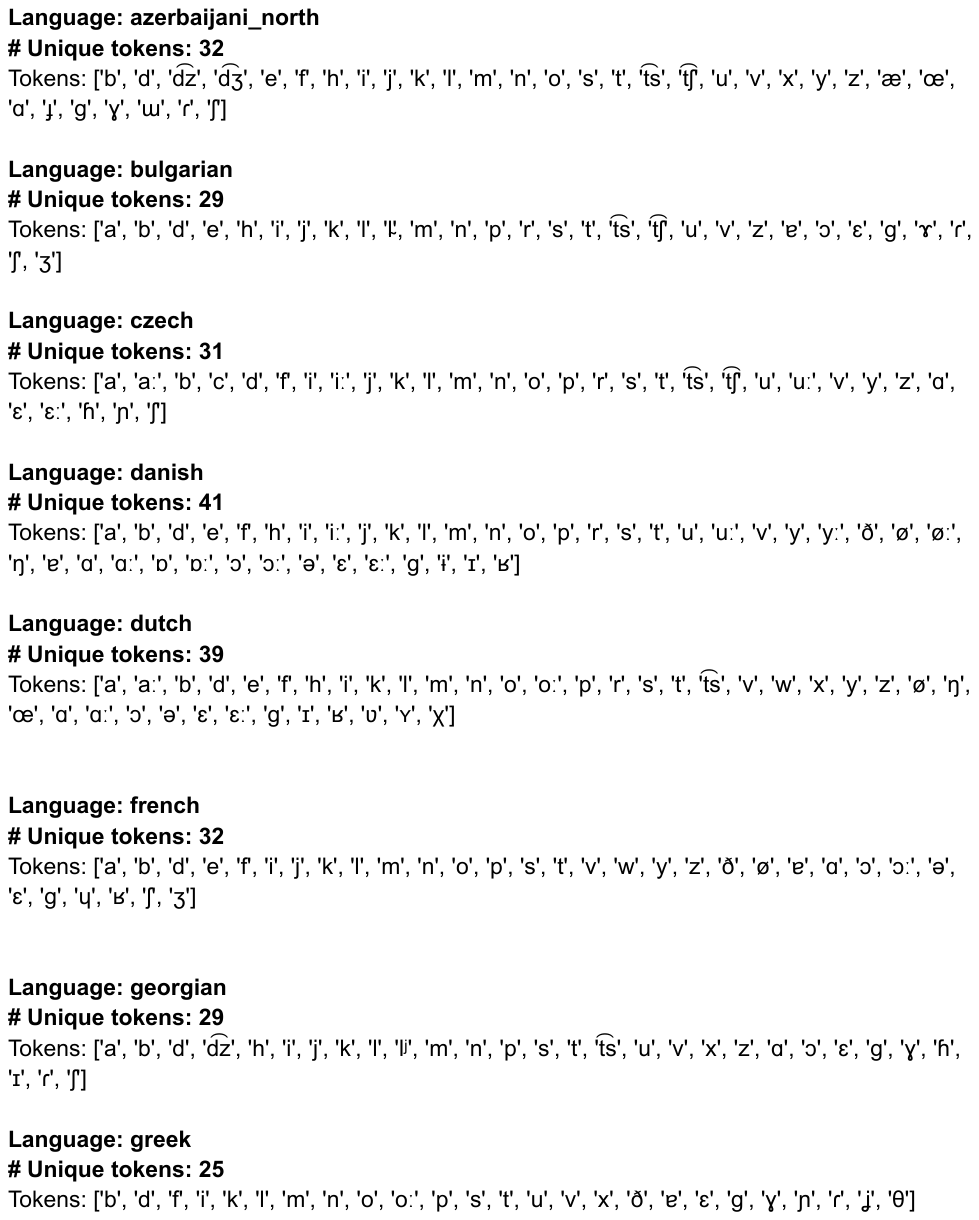}
\par\bigskip

\label{app:phon_cov} %

\clearpage
\begin{center}
\includegraphics[width=\textwidth, height=0.85\textheight, keepaspectratio]{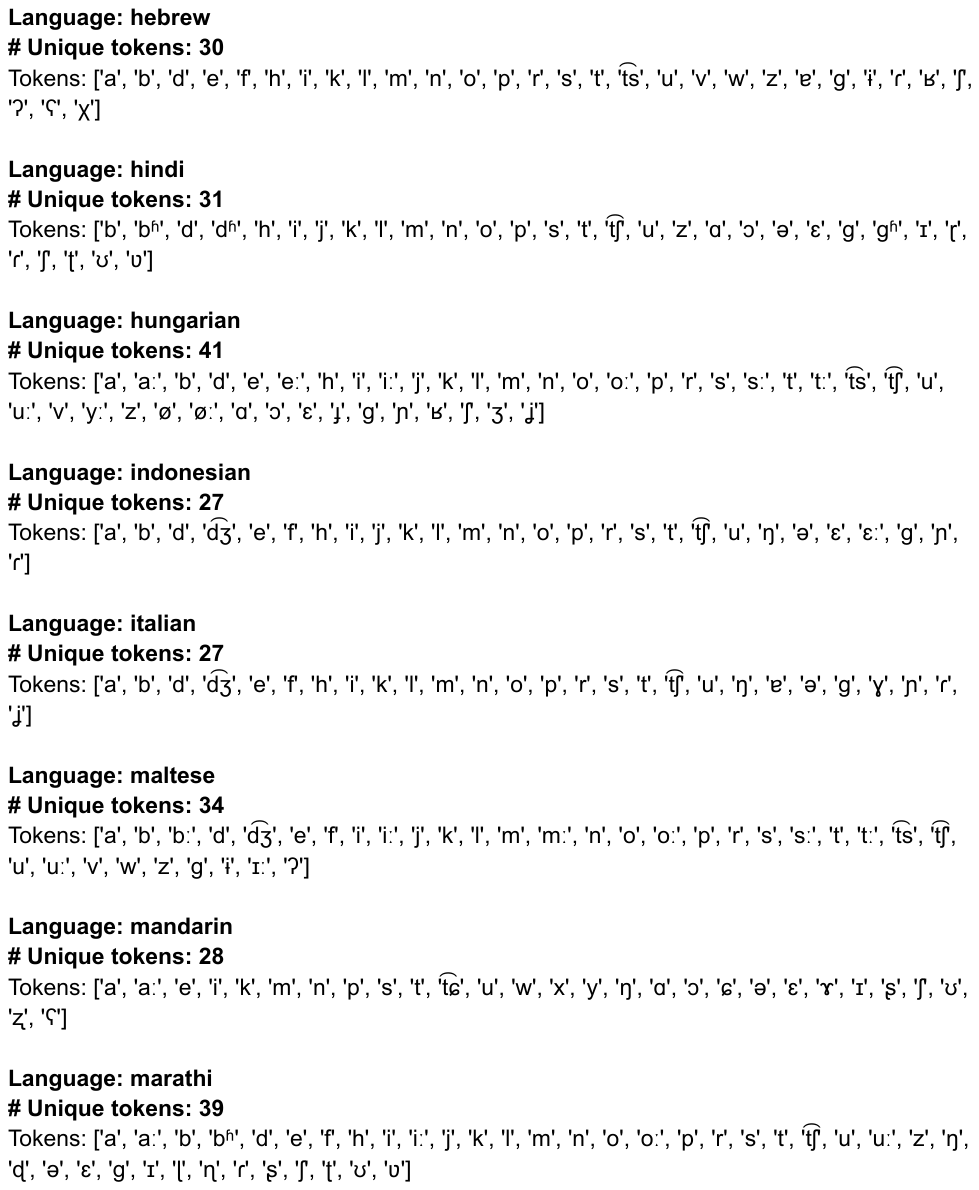}
\end{center}

\clearpage
\begin{center}
\includegraphics[width=\textwidth, height=0.85\textheight, keepaspectratio]{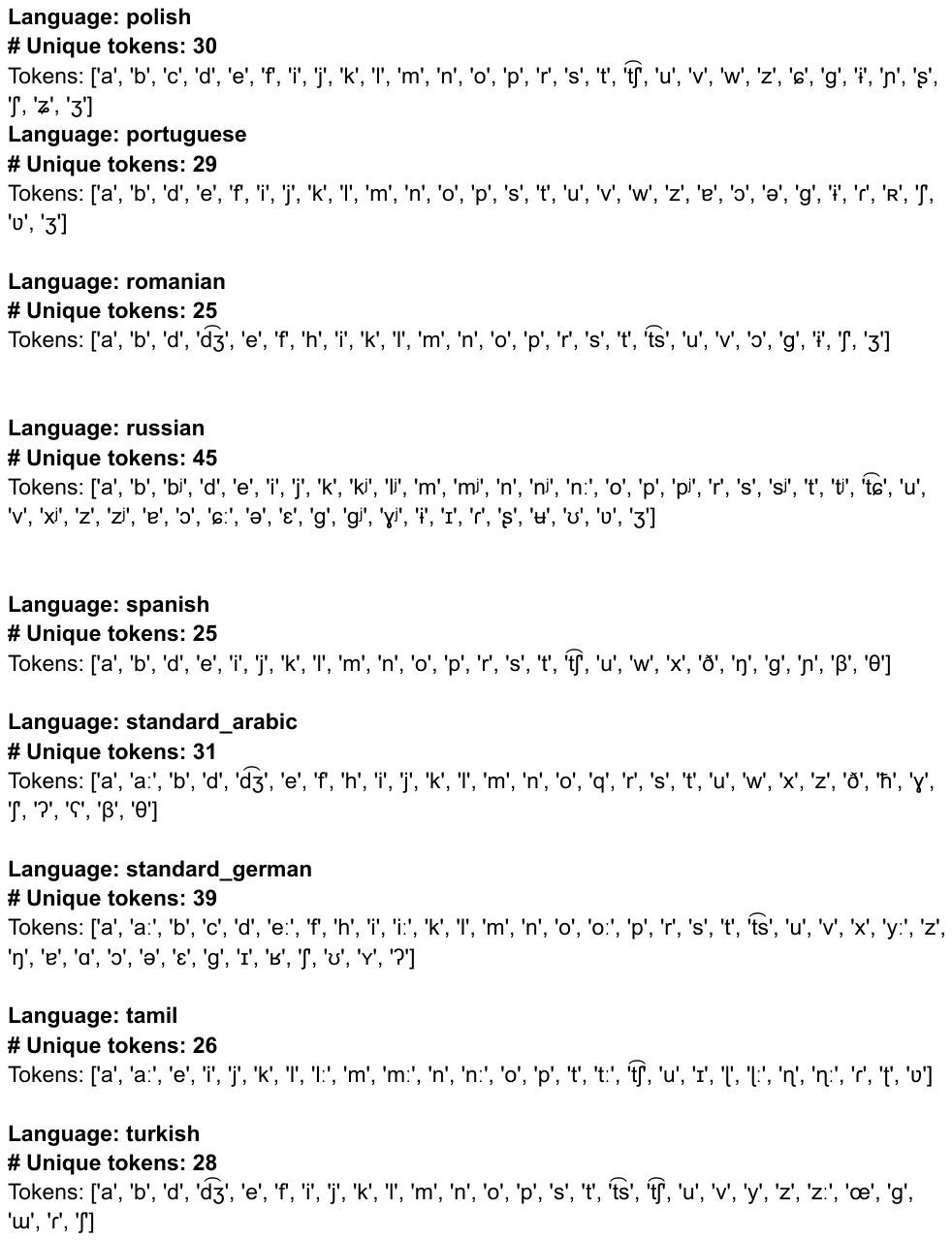}
\end{center}

\clearpage
\begin{center}
\includegraphics[width=\textwidth, height=0.85\textheight, keepaspectratio]{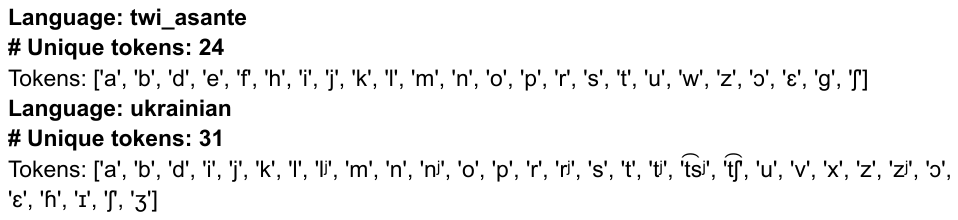}
\end{center}

\twocolumn
\end{document}